\documentclass[runningheads]{llncs}
\usepackage{graphicx}
\usepackage[hidelinks]{hyperref}
\usepackage{cleveref}[2012/02/15]
\crefformat{footnote}{#2\footnotemark[#1]#3}

\usepackage{framed,color,verbatim}
\definecolor{shadecolor}{rgb}{.9, .9, .9}

\newenvironment{code}%
   {\snugshade\verbatim}%
   {\endverbatim\endsnugshade}

\begin{document}
\title{WikiCausal: Corpus and Evaluation Framework for Causal Knowledge Graph Construction}
%
%
\author{Oktie Hassanzadeh}
%
\titlerunning{WikiCausal: Corpus and Evaluation Framework}
%
\institute{IBM Research \\
\email{hassanzadeh@us.ibm.com}
}
\maketitle              
\begin{abstract}
Recently, there has been an increasing interest in the construction of general-domain and domain-specific causal knowledge graphs. Such knowledge graphs enable reasoning for causal analysis and event prediction, and so have a range of applications across different domains. While great progress has been made toward automated construction of causal knowledge graphs, the evaluation of such solutions has either focused on low-level tasks (e.g., cause-effect phrase extraction) or on ad hoc evaluation data and small manual evaluations. In this Resource Track paper, we present a corpus, task, and evaluation framework for causal knowledge graph construction. Our corpus consists of Wikipedia articles for a collection of event-related concepts in Wikidata. The task is to extract causal relations between event concepts from the corpus. The evaluation is performed in part using existing causal relations in Wikidata to measure recall, and in part using Large Language Models to avoid the need for manual or crowd-sourced evaluation. We evaluate a pipeline for causal knowledge graph construction that relies on neural models for question answering and concept linking, and show how the corpus and the evaluation framework allow us to effectively find the right model for each task. The corpus and the evaluation framework are publicly available. 
\end{abstract}

\textbf{Resource Type}: Corpus \& Evaluation Framework 

\textbf{Corpus}: \url{https://doi.org/10.5281/zenodo.7897996}

\textbf{Evaluation Framework}: \url{https://doi.org/10.5281/zenodo.7902733}

\textbf{Data License}: Creative Commons Attribution 4.0 International

\textbf{Code License}: Apache 2.0

\section{Introduction}

Extracting and representing causal knowledge has been a topic of extensive research, with applications in decision support and event forecasting in a variety of domains such as sociopolitical event forecasting~\cite{HurriyetogluYMD21,SAGE:ijcai/MorstatterGSBAM19,embersAt4Years,RadinskyDM12_WWW}, enterprise risk management and finance~\cite{bromiley2015enterprise,iwamaHOPEGraphHypothesisEvaluation2021,Sohrabi19}, and healthcare~\cite{barnard-mayersAssessingKnowledgeAttitudes2021,prosperi2020causal,HealthData:IJSWIS.297145}.
One way to derive causal knowledge is by using observations in the form of structured data, and performing causal inference~\cite{pearlCausality2009,pearlCausalInference2010}. An alternative is to extract causal knowledge stated explicitly or implicitly in text documents. Such statements are abundant across domains and applications in various forms, such as analyst reports, news articles, financial reports, medical documents, books, and scientific literature. As a result, there is a body of research on extracting causal knowledge from text documents with the goal of turning the knowledge into structured form for various retrieval, analysis, and reasoning tasks.

To represent causal knowledge in a structured form, there is a body of work that structures the extracted causal knowledge as networks of cause-effect pairs, where each cause/effect is a phrase or a textual description. ATOMIC~\cite{ATOMIC} and CausalNet~\cite{LuoSZHW16CausalNet} are examples of such work, and both target applications in commonsense reasoning. CausalNet represents causal relations between simple terms (e.g., ``\textit{neuroma}" causes ``\textit{pain}") while ATOMIC represents causal relations between textual descriptions with variables (e.g. if ``\textit{X pays Y a compliment}'', then ``\textit{Y will likely return the compliment}'').  CauseNet~\cite{heindorf2020causenet} is a more recent example of such work that provides a comprehensive network of causal relations extracted from Wikipedia, where nodes are noun phrases. Recently, there has been a surge of interest in the use of causal relations in general-domain knowledge bases (e.g., Wikidata~\cite{VrandecicK14wikidata} and ConceptNet~\cite{DBLP:conf/aaai/SpeerCH17ConceptNet}), and in representing causal knowledge as a causal knowledge graph (KG)~\cite{DBLP:conf/semweb/Hassanzadeh21a,jaiminiCausalKGCausalKnowledge2022a}. Such representations further facilitate reasoning over the knowledge, e.g., for prediction~\cite{DBLP:conf/www/ShiraiBH23}. It also makes it possible to take advantage of knowledge graph completion methods to enrich the knowledge graph~\cite{DBLP:conf/semweb/KhatiwadaSSH22}.

Most applications relying on structured causal knowledge rely on a high level of accuracy (precision and recall) of the extracted knowledge. The NLP community has provided a variety of benchmarks for a range of related tasks, such as causal sentence classification~\cite{mariko-etal-2020-financial,mariko-etal-2021-financial,tan-etal-2022-event,caselli-vossen-2017-event-storyline}, cause-effect span identification~\cite{DunietzLC17,mirza-etal-2014-annotating-causal-timebank,prasadrashmiPennDiscourseTreebank2019}, and causal pair classification~\cite{HassanzadehBFSP20,hendrickx-etal-2010-semeval}, and unified benchmarks across the tasks~\cite{hosseiniCRESTCausalRelation2023,tanUniCausalUnifiedBenchmark2023}. On the other hand, there is no benchmark or unified evaluation framework for evaluating the quality of an extracted causal network or graph, and so prior work has mostly relied on crowd-sourced or human annotation. For example, ATOMIC~\cite{ATOMIC} is derived based on a crowd-sourcing framework and relies on inter-annotator agreement, while CauseNet~\cite{heindorf2020causenet} relies on estimating precision through a small human annotation and estimating recall through question answering.

In this paper, we present a dataset and an evaluation framework for assessing the quality of causal knowledge graphs extracted automatically from text documents. To the best of our knowledge, this is the first evaluation framework that allows for measuring the quality of end-to-end causal extraction solutions. Our target solutions are those that take textual corpora as input, and produce a knowledge graph of causal relations among a set of concepts. Our dataset is curated from event-related Wikipedia articles. The evaluation of recall is performed by measuring the coverage of causal relations that are already in Wikidata, as the majority of such relations are described in text in the associated Wikipedia articles. For the evaluation of precision, inspired by a recent trend in the use of large language models (LLMs) in lieu of crowdsourcing~\cite{he2023annollm,zhao-etal-2022-lmturk}, we devise a mechanism for automatically creating prompts and probing LLMs to measure the accuracy of the cause-effect concept pairs in the output that is being evaluated. To show the effectiveness of the evaluation framework, we use a modular causal knowledge extraction pipeline to generate four versions of a Wikidata-based causal knowledge graph, each using a different combination of pre-trained neural models. We show how the choice of models affects the quality of the output, and share some interesting lessons learned.

In what follows, we first describe the task. We then describe the process used to generate the evaluation corpus. In Section~\ref{sec:eval} we describe the methodology for evaluating precision and recall of the automatically generated causal knowledge graphs using the corpus. We present the details of our causal extraction pipeline and the results of our experiments. We end the paper with a discussion of some lessons learned from the experiments, and outline a few directions for future research. The dataset and the evaluation scripts are made publicly available.


\section{Task Definition and Use Cases}
\label{sec:definition}

Our target task is as follows: given a corpus of text documents and a select set of concepts (e.g., from an existing knowledge graph), automatically generate a causal knowledge graph in which nodes are the given concepts, and an edge between two concepts indicates a causal relation between the concepts. The select concepts in the knowledge graphs could be either event-related classes, or instances of such classes. The class concepts could belong to an ontology, with a class hierarchy. We assume that no  annotations or training data are available. That is, while we know what concept each document is associated with, we do not have annotations of concepts or relations in the corpus.

\begin{figure}[t]
\begin{center}
    \includegraphics[width=0.98\textwidth]{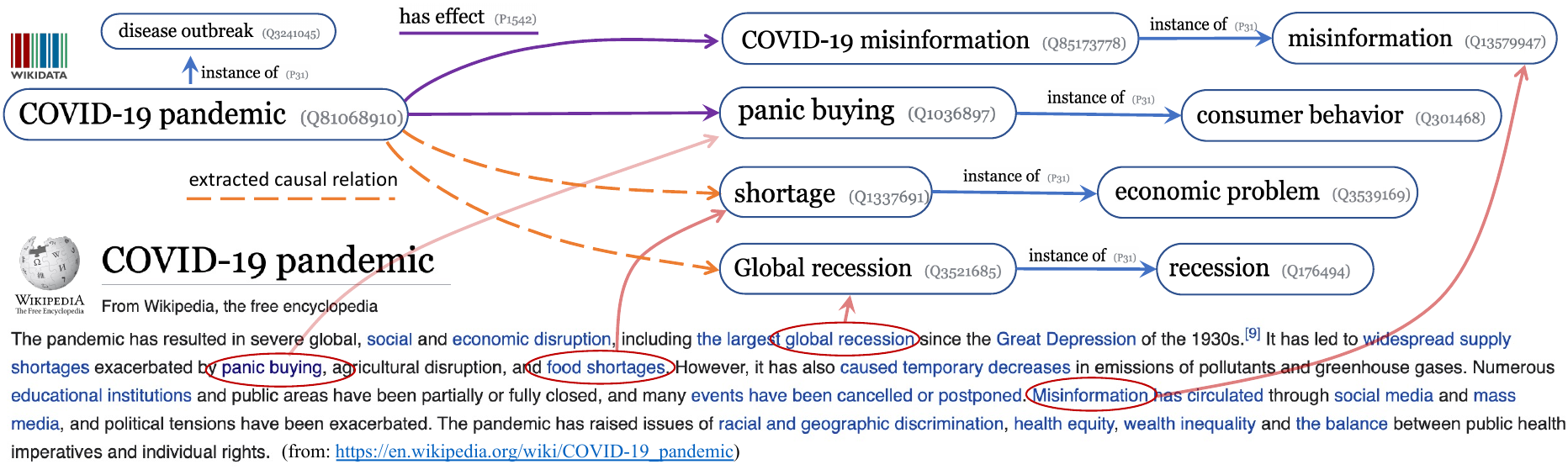}
    \caption{Examples of Event-Related Causal Knowledge in Wikidata and Wikipedia}
    \label{fig:examples}
\end{center}
\end{figure}

Figure~\ref{fig:examples} shows a snippet of a document from our Wikipedia-based corpus, along with a set of concepts from Wikidata. As we can see in this example, Wikidata contains a few causal relations for the concept ``COVID-19 pandemic", all of which are described also in the text of its English Wikipedia article. On the other hand, there are a number of causal relations described in text that are not on Wikidata. For this example, an application of the task defined above is to augment and/or validate the available causal knowledge. This case can also arise in applications such as healthcare or enterprise risk management, where part of the causal knowledge has already been captured in a structured form.

Another use case for this task is construction of a domain-specific causal knowledge graph from a given corpus. For example, an analyst or a chief risk officer in an organization in a certain industry may want to study the impact of certain economic events (e.g., a change in employment rate, recession, technology trends, or natural disasters) on certain events of interest for that industry. In such cases, there is often a wealth of knowledge around these topics in internal analyst reports or external records of similar organizations (e.g., from financial statements and annual reports of public companies). Turning such knowledge into a structured form will facilitate deeper retrieval and analysis, and enable the application of automated planning and risk management solutions~\cite{Sohrabi19,ijcai2022p850neatToolkit}.

\section{Corpus Creation}
\label{sec:corpus}
\begin{figure}[t]
\begin{center}
    \includegraphics[width=0.98\textwidth]{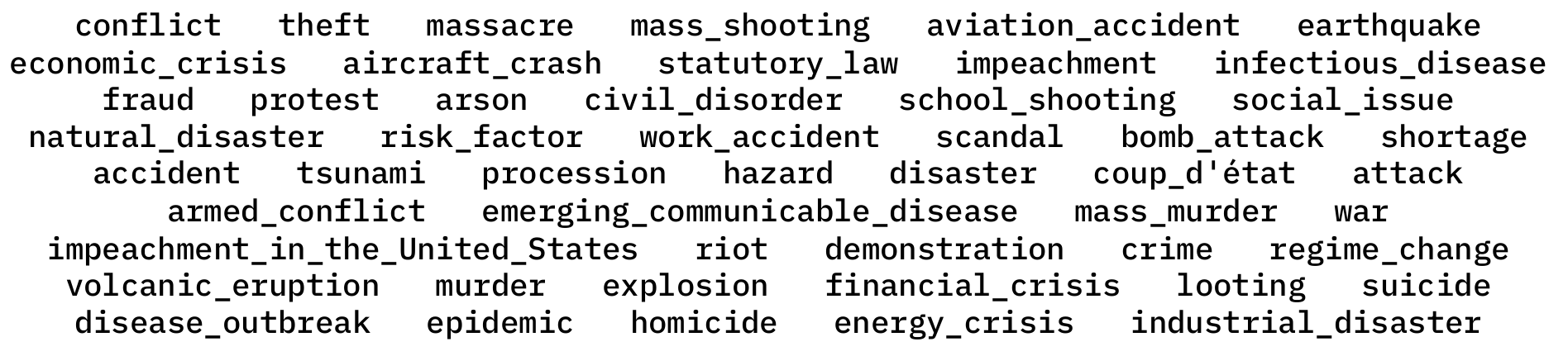}
    \caption{Selected Top-Level News Event Concepts}
    \label{fig:event_concepts}
\end{center}
\end{figure}

Given our task definition, we curate a collection of text documents, each associated with an event-related concept. We use Wikipedia as the source of our text documents and Wikidata as our source of event-related concepts. The first step in curating our corpus is identifying a set of event-related concepts in Wikidata. We do so by querying Wikidata for concepts that have associated Wikinews articles. An associated Wikinews article implies that the article's topic is on a newsworthy event instance. We then find the set of all the classes of the retrieved instances that are subclasses of class \texttt{{\it occurrence (Q1190554)}} to ensure that the chosen class is an event class as some non-event classes also have links to Wikinews. We then further manually verify each of the concepts and drop those that are not event-related. We do so since currently some non-event related concepts are (possibly erroneously) sub-classes of \texttt{occurrence (Q1190554)}. For the first version of our corpus, this results in a select set of 50 top-level event-related concepts in Wikidata. Figure~\ref{fig:event_concepts} shows the labels of these concepts.

\begin{figure}[t]
\begin{center}
    \includegraphics[width=0.98\textwidth]{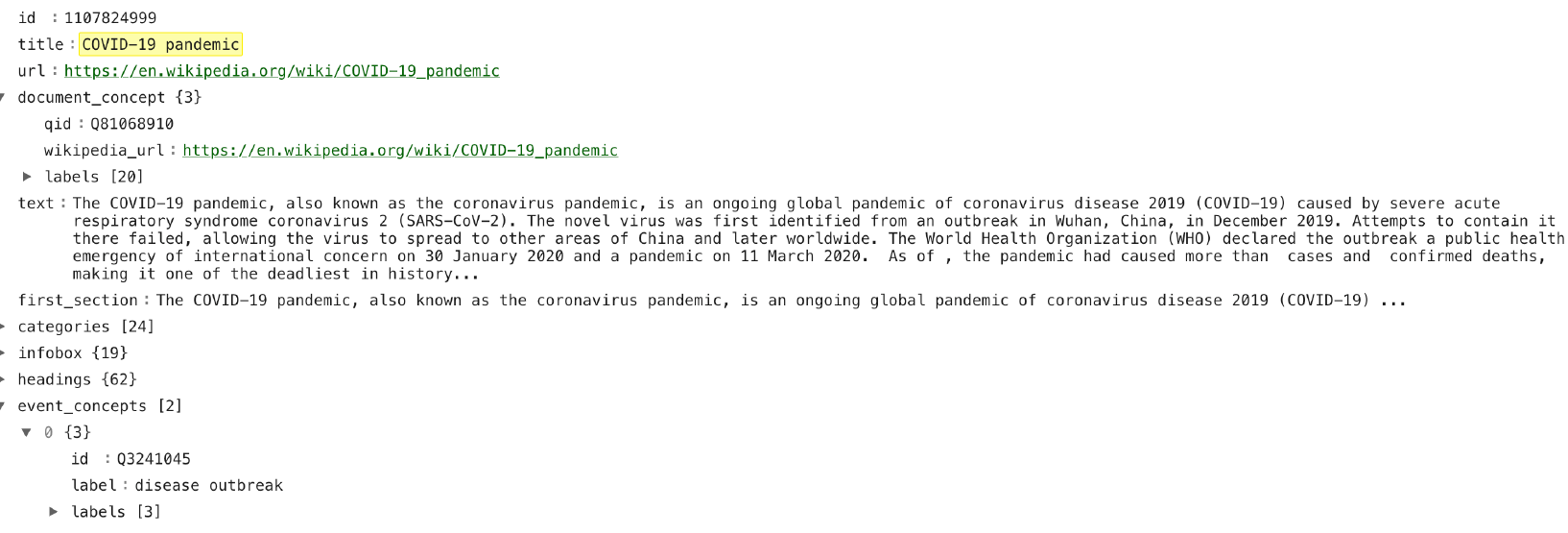}
    \caption{Example JSON Document}
    \label{fig:example_document}
\end{center}
\end{figure}

The next step is to retrieve all the instances of the identified event-related classes in Wikidata. We then use the Wikipedia ``sitelinks" to collect the URL of all the associated English Wikipedia documents. We use the list of URLs over a dump of English Wikipedia to retrieve the associated Wikipedia articles, and process the contents of each article into plain text in addition to some meta-data about the page such as section headlines, categories, and infoboxes. We store the outcome in the form of a \texttt{jsonl} file, with each line being a JSON object containing the page contents, meta-data, and associated event concept(s). Figure~\ref{fig:example_document} shows an example of a JSON document related to the example shown in Figure~\ref{fig:examples}. The fields in each JSON object are:
\begin{itemize}
    \item \texttt{id}: Wikipedia page identifier.
    \item \texttt{title}: Wikipedia page title.
    \item \texttt{url}: Wikipedia page URL.
    \item \texttt{document\_concept}: The Wikidata concept (instance) associated with the document. It comes with the \texttt{QID} and all the labels for the concept, which as will see in Section~\ref{sec:experiments} can be used for causal relation extraction.
    \item \texttt{text}: This is the field that contains the full clean text contents of the Wikipedia article, to be used for causal knowledge extraction.
    \item \texttt{first\_section}: A separate field for the first section of the article, as it often contains a summary with all the key causal knowledge. Less scalable methods can use only this field for extraction.
    \item \texttt{categories}: List of categories for the page. Categories can also be useful in identifying the topics covered in a page, which can be useful for the extraction process.
    \item \texttt{infobox}: structured infobox fields and values.
    \item \texttt{headings}: section headings of the Wikipedia page.
    \item \texttt{event\_concepts}: The set of top-level event concepts (classes) associated with the page. These are seed event concepts that are superclasses of the \texttt{document\_concept}.
    \item \texttt{timelines}: Some Wikipedia pages have a timeline section describing sequences of sub-events that occurred during the described event. While not the focus of the evaluation in this paper, such sequences can be mined for causal knowledge, e.g. using event sequence models capable of handling noisy ordered event sequences~\cite{DBLP:conf/ijcai/BhattacharjyaSH22}.
\end{itemize}

The first version of the dataset is generated from September 1, 2022 dump of Wikipedia and contains 68,391 articles, including 65,358 pages with textual contents and 3,216 redirect pages that are linked from some Wikidata concepts and facilitate the discovery of the textual contents for each Wikidata concept. There are 63,634 unique \texttt{document\_concept}s which means a number of Wikidata concepts have more than one page associated with them. 
The average length of \texttt{text} is 9,245 characters and the average length of \texttt{first\_section} is 905 characters. Each article is associated with an average of 2.2 \texttt{event\_concepts}. 

\section{Evaluation Framework}
\label{sec:eval}

As with any automated knowledge graph construction task, we need to measure the quality of the output both in terms of the number of causal relations expressed in text that have been extracted (recall) and the number of extracted causal relations that are accurate (precision).

\subsection{Recall Evaluation}
Given that manually extracting all the expressed causal relations over the corpus is not feasible, our automated recall evaluation relies on existing causal relations in Wikidata. Although the number of causal relations in Wikidata is limited and much less than the available causal knowledge expressed in Wikipedia documents, the majority of such relations are described in Wikipedia articles, and so measuring the ability of an automated method to discover those relations provides us with a good estimate of the actual recall of the method. Over time, we expect the high-confident accurate causal relations to be added to Wikidata, and so this evaluation strategy can become an even more reliable measure of recall. In our experiments with causal knowledge extraction methods, we discovered that different methods perform differently with respect to extracting relations between instances and classes. As a result, we apply our accuracy measures separately for relations that include at least one instance, and those that are between classes only.

For recall evaluation, we first construct a causal knowledge graph from the existing causal relations in Wikidata and our selected seed concepts and all their instances. We refer to this graph as the ``Base KG". Our recall evaluation script takes the Base KG and the output of causal extraction as inputs, and reports the following measures:
\begin{itemize}
    \item \texttt{recall} The ratio of causal relations in the Base KG that can be found in the extraction output.
    \item \texttt{hit\_count} The number overlapping causal relations between the Base KG and the extraction output.
    \item \texttt{rel\_count} The number of extracted causal relations.
    \item \texttt{base\_kg\_size}: The number of causal relations in the Base KG.
    \item \texttt{base\_count}: The number of unique concepts from the Base KG that is in the output. 
    \item \texttt{base\_coverage}: The percentage of unique concepts in Base KG that can be found in the output.
\end{itemize}

The above measures are calculated once for the full output and Base KG, once for the portion of the output and Base KG covering only classes, and once for the portion of the output and the BaseKG relations that include at least one instance event.

\subsection{Precision Evaluation}

In the absence of a complete knowledge graph for a given corpus, the standard way to evaluate the precision of the extracted knowledge is manual evaluation. Manual evaluation, however, is tedious and time-consuming, which limits the possibility of experimenting on a large scale with a wide range of methods and parameters. Inspired by a recent trend in the use of large language models (LLMs) as an alternative to crowd-sourcing and manual annotation~\cite{he2023annollm,zhao-etal-2022-lmturk,zheng2023judging}, we devise a mechanism to automatically create prompts for generative LLMs to evaluate the precision of the extracted causal relations. This approach works well for our corpus and task since LLMs have been exposed to the knowledge that is available on Wikipedia and Wikidata and are therefore likely to perform very well in the verification of the extracted relations. It is important to note that we do not use LLMs as a causal extraction method since our goal is the evaluation of generic extraction methods that can handle proprietary sources of knowledge that are less likely to be in the sources that LLMs are trained on. Our goal in this paper is not to curate a large-scale Wikipedia-based causal KG but to evaluate generic causal extraction methods. As mentioned earlier, such methods have a range of applications in risk management, intelligence analysis, and healthcare, where there is access to proprietary sources of knowledge.

Our precision evaluation script takes as input a causal knowledge graph consisting of cause-effect pairs of concepts and uses each pair to generate a prompt for a generative LLM to verify the accuracy of the extracted pair. There are many ways to use LLMs for this validation task, and for each approach, there are several available models and parameters to be chosen. In order to find the best approach, we use the Base KG relations to measure the performance of each approach. This is similar to asking verification questions to crowd workers to find high-performing workers and drop the low-performing ones. We observed in our experiments that instruction-tuned generative LLMs perform better at the task, as verified using our Base KG.
As an example, an extracted pair \texttt{(cause, effect)} can be turned into the following prompt:
\begin{code}
Definition: Answer the question with a yes or no.
Now complete the following example -
Input: Question: Could {cause} result in {effect}? 
Output:
\end{code}
\noindent The generative model then provides a \texttt{yes} or \texttt{no} answer to the question. We can provide the same prompt several times, and return an output that is in the majority of the outputs. This is similar to asking the same question to  several crowd workers and returning the answer with the most inter-annotator agreement. It is also possible to provide a confidence score for each answer. Here again, the Base KG is used to find the best-performing model and parameters.

As with recall evaluation, we observed that different methods perform differently for class-level causal relations and instance-level relations. Most LLM-based verification strategies also perform better over class-level causal relations, as they result in more generic prompts that such models can more consistently handle. As a result, we report precision scores for instance-level relations, class-level relations, as well as the full output.

\section{Experiments}
\label{sec:experiments}

In this section, we report our preliminary results of using the corpus and our evaluation framework to evaluate a causal knowledge extraction pipeline that relies on the extraction of cause-effect phrases and linking the outcome to event concepts. As a part of our evaluation framework, in addition to the evaluation scripts and the corpus, we have made the extracted knowledge graph outputs and the results on these outputs publicly available.

\subsection{Causal Extraction Pipeline}

\begin{figure}[t]
\begin{center}
    \includegraphics[width=0.98\textwidth]{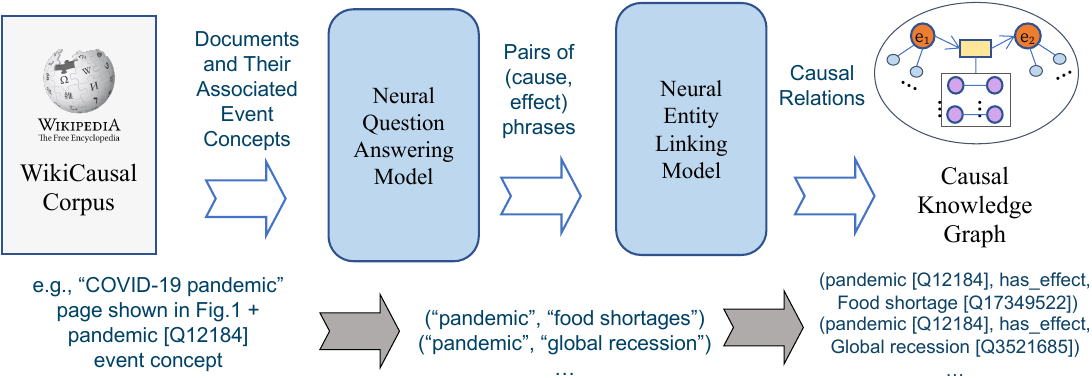}
    \caption{Question Answering and Linking (QAL) Pipeline for Causal Knowledge Extraction}
    \label{fig:pipeline}
\end{center}
\end{figure}

The causal extraction pipeline we use in our experiments is in part based on our previous work~\cite{DBLP:conf/semweb/Hassanzadeh21a}, that splits the causal extraction process into two steps: 1) extraction of pairs of cause-effect phrases 2) linking each cause and effect to event concepts. The pipeline is depicted in Figure~\ref{fig:pipeline}. As mentioned earlier, there is a range of methods for cause-effect pair extraction, including supervised sequence tagging methods~(e.g., \cite{tacl/DunietzLC17}) as well as weakly-supervised and unsupervised methods (e.g., ~\cite{BhandariFHSS21,DBLP:conf/case-ws/SahaGNHYS22CausalNews} and references therein). For two of the four knowledge graphs used in our experiments, we use the causal pairs from CauseNet~\cite{heindorf2020causenet} which is based on the application of a number of state-of-the-art extraction methods including a supervised sequence tagger. The other two knowledge graphs rely on question answering using the associated Wikidata concept labels in the corpus as seed terms to extract causal relations. We refer to these KGs as QAL (Question Answering \& Linking) KGs, which are constructed by first using the seed terms to create open-ended causal questions. For example, a seed term \texttt{event} can turn into a question ``What does \texttt{event} lead to?" or ``What causes \texttt{event}?". The question is then asked from either the opening paragraph or all the paragraphs in the associated articles. The answer span along with the label used in the input result in a cause-effect pair, with one side already linked to an event concept in the seed set.

Once we have a collection of cause-effect pairs of phrases, the next step is to link them to event concepts. While this task is similar to the classic entity linking tasks studied extensively in the literature, most existing solutions are designed to handle named entities (e.g., persons, organizations) and not event concepts. What we need is a generic event disambiguation method that takes a mention and a context as input and returns the identified event concepts as output. For the results reported in this paper, we use the entity disambiguation function of BLINK~\cite{wu2019zero}, using the cause or effect phrase as span and the phrase used for extraction of the pair as the context for disambiguation. In the future, we will experiment with other alternatives, such as an adaptation of BLINK called EVELINK~\cite{yuEventLinkingGrounding2023EveLINK} that is tuned to perform significantly better than BLINK for linking event mentions.

\subsection{Evaluation Framework Repository and Settings}

\begin{figure}[t]
\begin{center}
    \includegraphics[width=0.98\textwidth]{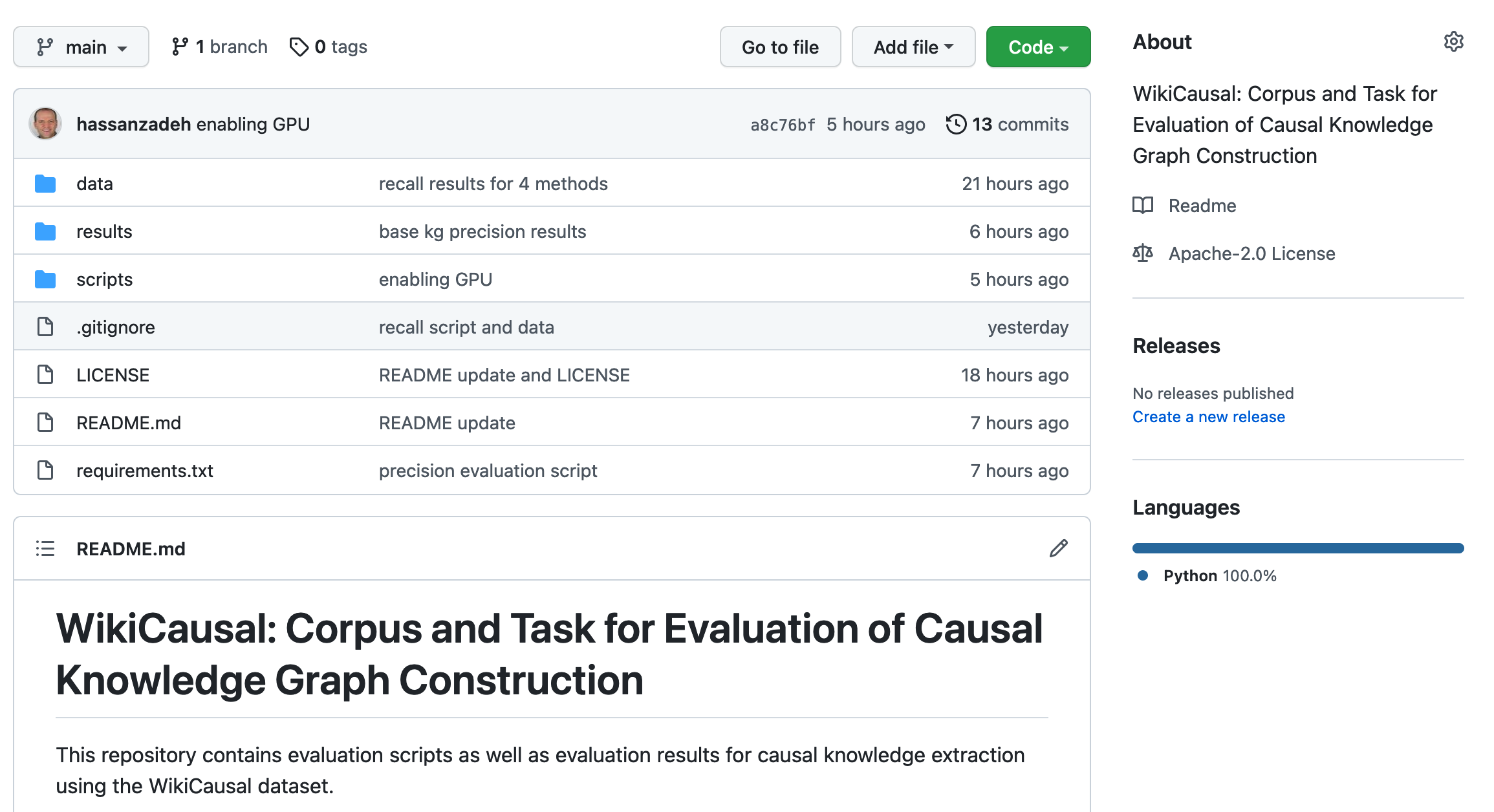}
    \caption{Screenshot of GitHub Repository of WikiCausal Evaluation Framework\cref{footnote2}}
    \label{fig:repo}
\end{center}
\end{figure}

For reproducibility, all the results reported in this section are obtained using scripts and data that are released publicly as a snapshot of our git repository. A screenshot of the repository is shown in Figure~\ref{fig:repo}\footnote{\label{footnote2}The screenshot is included to make this paper self-contained for ISWC review. URL: \url{https://github.com/IBM/wikicausal/}}. We have made our git repository public so that future improvements (by our team and the community) as well as future results can similarly be shared, and our results page can act as a public leaderboard for the latest results on the corpus.

For the results reported in this paper, we have used version 1 of our released corpus which is derived from a September 1, 2022 dump of English Wikipedia. We experimented with a number of LLMs for our precision evaluation including large and proprietary models, and ended up picking \texttt{allenai/tk-instruct-3b-def} as a publicly available model that can run without GPUs (albeit slow) so the evaluation can be performed without requiring hard-to-obtain resources. Each precision evaluation takes about an hour to run using CPUs only, and a few minutes with one V100 GPU with 16GB of memory. Since the model is instruction-tuned, the prompts are in the form of instructions as shown in the previous section.

\subsection{Results}

\begin{figure}[t]
\begin{center}
    \includegraphics[width=0.98\textwidth]{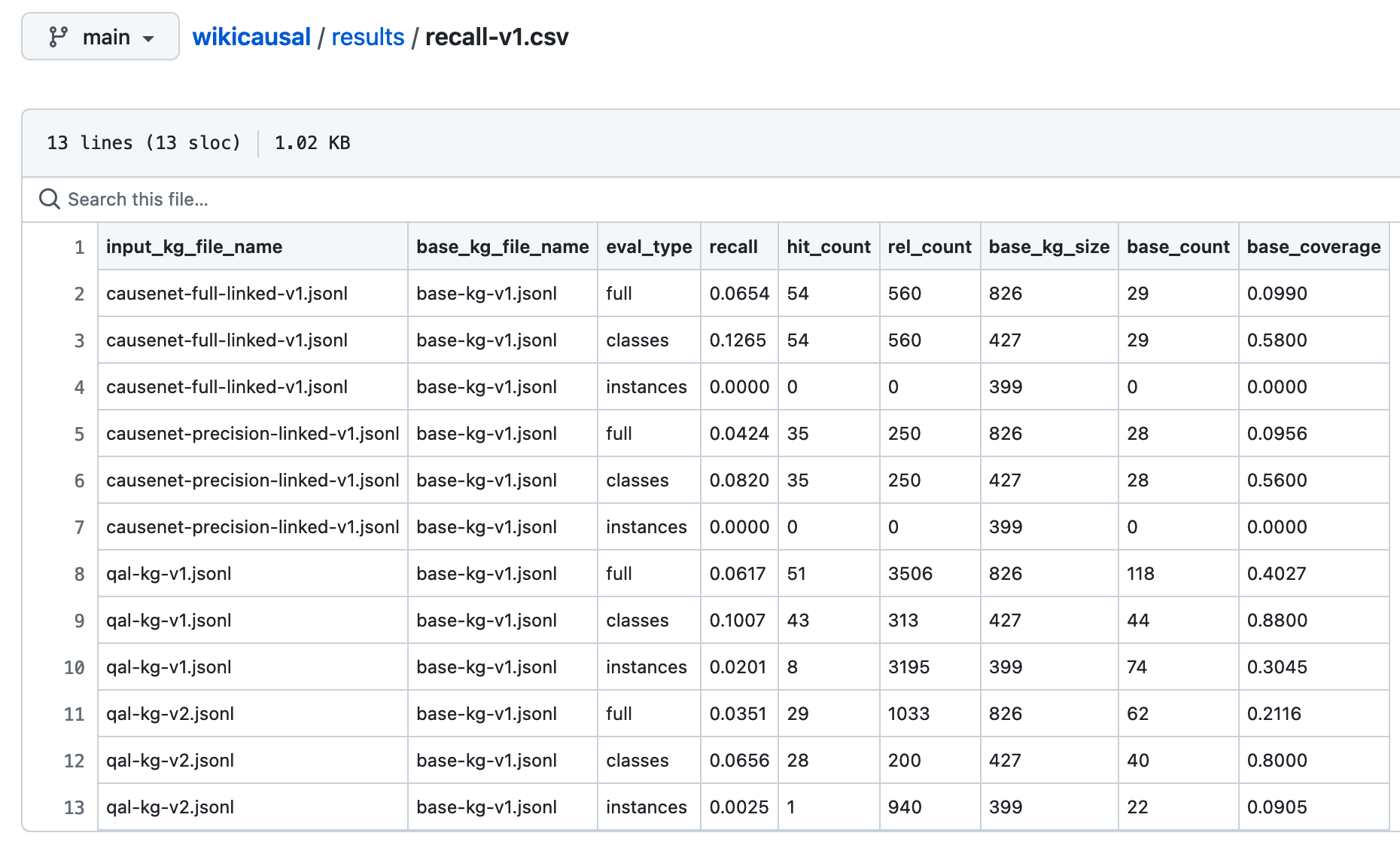}
    \caption{Recall Evaluation Results (v1) - Table View\cref{footnote4}}
    \label{fig:recall}
\end{center}
\end{figure}

\begin{figure}[t]
\begin{center}
    \includegraphics[width=0.98\textwidth]{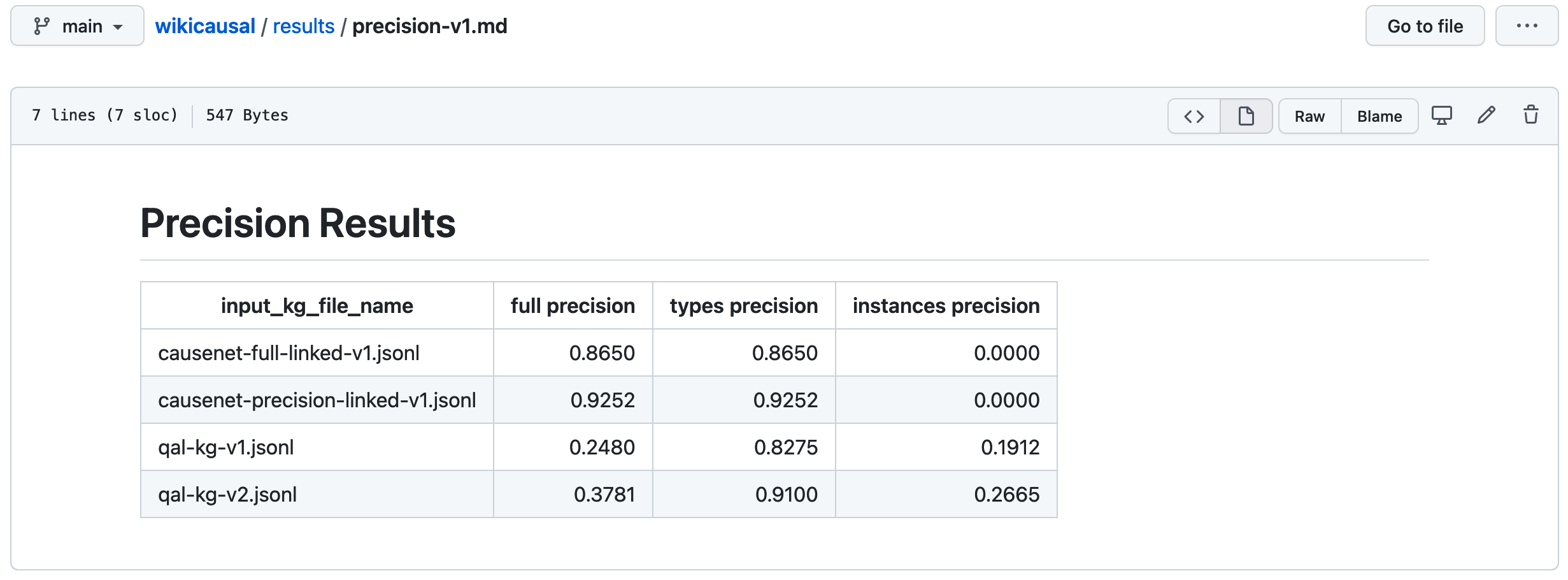}
    \caption{Precision Evaluation Results (v1) - Markdown View\cref{footnote5}}
    \label{fig:precision}
\end{center}
\end{figure}

Figures~\ref{fig:recall} and \ref{fig:precision} show the results of recall and precision evaluation over four extracted knowledge graphs:
\begin{itemize}
    \item \texttt{causenet-full-linked-v1}: The CauseNet Full data~\cite{heindorf2020causenet}, turned into a Causal KG through concept linking using BLINK~\cite{wu2019zero}.
    \item \texttt{causenet-precision-linked-v1}: The CauseNet Precision subset (publicly-available higher-precision subset of CauseNet), also turned into a KG through linking using BLINK.
    \item \texttt{qal-kg-v1}: Question Answering \& Linking (QAL) pipeline that uses DistilBERT~\cite{DBLP:journals/corr/abs-1910-01108DistilBERT} fine-tuned on SQuAD2.0~\cite{squad2} dataset, with extractions only on the \texttt{first\_section} of the articles in the corpus. The pairs are then linked to event concepts using BLINK. 
    \item \texttt{qal-kg-v2}: QAL pipeline that uses \texttt{mfeb/albert-xxlarge-v2-squad2} model\footnote{\url{https://huggingface.co/mfeb/albert-xxlarge-v2-squad2}} also fined-tuned on SQuAD2.0, over all the \texttt{text} contents of the articles in the corpus. Again, the output is linked using BLINK.
\end{itemize}

Recall evaluation results in Figure~\ref{fig:recall}\footnote{\label{footnote4}Screenshot shown instead of the table since this Resource track paper aims at describing the publicly available resource, which includes the results table as shown. A comprehensive evaluation is not the goal of this paper. URL: \url{https://github.com/IBM/wikicausal/blob/main/results/recall-v1.csv}} provide some interesting insights on how the KGs compare in terms of their ability to extract causal relations that can be found on Wikidata. Our first observation is that all the absolute recall numbers are very low, due to the fact that: a) the supplied text may not contain all the causal knowledge in the ground truth as Wikidata relations are not always described in their associated Wikipedia articls, b) causal relations are described in a variety of implicit and explicit ways, making automated extraction very challenging. Another observation is that all the methods we have examined perform significantly better at extracting class-level relations. CauseNet Full finds 54 of the 427 class-level causal relations, although it only covers 29 of the 50 event classes in the corpus. The best-performing QAL KG covers 43 class-level relations, including 44 of the 50 event classes in the corpus. CauseNet does not extract any instance-level causal relations, which is expected given the way CauseNet extraction pipeline is designed. The QAL pipelines find a very small number of instance-level relations, with \texttt{qal-kg-v1} including 8 such relations.

Precision evaluation results are shown in Figure~\ref{fig:precision}\footnote{\label{footnote5}URL: \url{https://github.com/IBM/wikicausal/blob/main/results/precision-v1.md}}. Precision score of CauseNet Precision is very close to the estimate of precision provided by Heindorf et al.\cite{heindorf2020causenet}, with a precision of 92.5\% vs. the full subset having a precision of 86.5\%. Among the QAL KGs, \texttt{qal-kg-v2} which has a lower recall has a much higher precision, showing the classic precision-recall trade-off but also proving \texttt{qal-kg-v2}'s models to be more effective in use cases that rely on high precision.

\section{Discussion, Lessons Learned, and Future Work}
\label{sec:discussion}

\paragraph{Task Difficulty \& Use Cases}
As our recall and precision results show, the defined task is very challenging for pipelines built using state-of-the-art neural models for knowledge extraction. However, the high level of precision means that the application of these methods would result in the addition of a significant number of causal relations to the base knowledge graph. Such discovered relations can play a critical role in end applications such as event prediction~\cite{ShiraiBH23}. We also observed that our automated precision evaluation method underestimates the actual level of precision for instance-level relations due to the use of longer labels that are also less frequently observed in the source that large language models are trained on. Still, the precision evaluation method provides a reliable way of comparing different methods' level of precision. In our manual inspection of the results, we see many highly-interesting and causal relations that are difficult to extract using classic rule-based methods. Finally, it is important to note that many relations in the ground truth used for recall evaluation may not be described in the associate article, and so the recall metric should be considered mainly for comparison of different methods and not for measure the absolute recall of the methods.

\paragraph{Sustainability}
As the current snapshot of our evaluation framework repository shows, we are making all the data and scripts that were used for the results in this paper public. We also plan to make it easy for the community to contribute new outputs, results, and extensions to the evaluation scripts and to the corpus. Our github repository (shown in Figure~\ref{fig:repo} and included in the Zenodo snapshot) is now publicly available at \url{https://github.com/IBM/wikicausal}. We are currently working on release version 2 of the corpus. While it is essential to fix the document corpus for maintaining a public leaderboard of results, we will extend the code base to include corpus generation to allow the community to enhance the quality of the corpus and work on generating future versions of the corpus.

\paragraph{Causal Knowledge Extraction Challenges}
Our results so far shed light on a number of challenges in automated extraction of causal knowledge from textual corpora. One challenge highlighted by our results so far is related to the ability of the examined methods in extraction of causal relations between event instances. Another challenge we observed is related to the linking of cause or effect phrases to event-related concepts in Wikidata. In our manual inspection of the results, we observe that our BLINK-based linking method is the source of many inaccurate relations in the extracted KGs. We are currently exploring the use of recent work on event linking~\cite{yuEventLinkingGrounding2023EveLINK}. One challenge in use of BLINK and EVELINK disambiguation methods is the resource-intensive nature of such solutions, which makes the overall pipeline prohibitively slow especially given the number of different models and parameters that need to be evaluated.

\paragraph{Impact: Beyond Causal Knowledge}
Our results so far clearly call for more work on automated causal knowledge extraction methods. We believe with more accurate and scalable methods, we will start to observe a major improvement in the performance of solutions that rely on the availability of such knowledge (e.g.,~\cite{ijcai2022p850neatToolkit,iwamaHOPEGraphHypothesisEvaluation2021,DBLP:conf/www/ShiraiBH23,Sohrabi19}). Future work should also examine the use of generic knowledge extraction methods (e.g., IntKB~\cite{kratzwald-etal-2020-intkb}) over this corpus and adapt or extend such solutions to perform better in the extraction of causal knowledge. We believe our corpus can also be used as a means of evaluating certain flavors of knowledge extraction methods, such as methods relying on distant supervision~\cite{mintz2009distant}, and ensemble methods combining the results of rule-based, supervised, and unsupervised knowledge extraction methods.

\paragraph{Comprehensive Evaluation of Causal Knowledge Extraction Methods} Since the goal of this Resource Track paper is to present a resource, the experiments described in Section~\ref{sec:experiments} aim to show the usefulness of the resource (the corpus and the evaluation framework) in evaluation of causal knowledge extraction methods. It is important to note that state-of-the-art knowledge extraction methods are currently primarily based on much larger and resource-intensive LLMs such as GPT-4. While future work includes an evaluation of the use of such models along with prompt engineering techniques and RAG~\cite{lewis2020retrieval}, which will very likely yield superior results comparing with the methods we have investigated in this paper, one should consider the fact that such models are exposed to not only Wikipedia articles and so the WikiCausal corpus, but many other sources that include causal knowledge about the generic events that can be found on Wikidata.
We believe the primary use case of WikiCausal is the evaluation of domain-specific models trained and/or fine-tuned on a corpus similar in size and characteristics to WikiCausal. This is a very practical use case, where an enterprise or an organization needs to curate a high-quality causal knowledge graph purely based on an available corpus and collection of event concepts of interest. As stated in Section~\ref{sec:definition}, such structured KGs can play a critical role in decision support applications.

\paragraph{Extensions}
As stated earlier, our corpus has fields that can be used for different kinds of knowledge extraction methods. In this paper, we have only used the text field for evaluation of methods of knowledge extraction from text. Future work can take advantage of the more structured fields such as timelines, lists, and infoboxes. Another major limitation of our corpus is that it consists of English Wikipedia articles only. Given the multilingual nature of Wikidata and Wikipedia, an interesting avenue for future research is extending the corpus to a multilingual corpora, applying multilingual models, and leveraging extractions across languages to create a more comprehensive causal knowledge graph.

\section{Conclusion}
\label{sec:conclusion}
We presented a corpus, task, and evaluation framework for extracting causal knowledge from textual corpora. Using our evaluation framework, we presented the results of our evaluation of four different automatically extracted knowledge graphs of causal relations. As part of our evaluation framework, we developed a method for measuring the relative recall of various extraction methods using Wikidata's existing causal knowledge. In lieu of crowdsourcing or manual evaluation, we devised a novel method for gauging the precision of extraction methods by employing large generative language models. We have made our corpus and evaluation framework permanently available to the public, and we intend to establish a community and leaderboard for the task and for future extensions of the corpus and evaluation framework. While our work targets an important set of causal knowledge applications for reasoning and prediction, it also fills an important gap in the literature on resources for the evaluation of knowledge extraction methods from text.

\paragraph*{Resource Availability Statement:} WikiCausal code base is available at \url{https://github.com/IBM/wikicausal}. The corpus is permanently available at \url{https://doi.org/10.5281/zenodo.7897996}. A permanent snapshot of the github repo is available at \url{https://doi.org/10.5281/zenodo.7902733}. The code base license as specified is Apache 2.0, and the data/corpus license is Creative Commons Attribution 4.0 International since it is derived from Wikipedia. The code base includes a README with installation and usage instructions. Sustainability plan is discussed in the paper. We have made and will continue to make every effort to ensure that the answer to the majority of the questions on the \href{https://iswc2024.semanticweb.org/event/3715c6fc-e2d7-47eb-8c01-5fe4ac589a52/websitePage:cb52690e-aefa-4d99-8e16-f354276bd7df?tm=QLST-3OD03RslebCRIVzGyOABTZJAKLJIULY8p37Rms}{Resource Track Call For Papers} page as per criteria for ``Impact", ``Reusability", ``Design \& Technical Quality", and ``Availability" is positive.

\paragraph*{Acknowledgements} The author would like to sincerely thank Mark Feblowitz for his contributions to this resource and his assistance in running some of the experiments described in Section~\ref{sec:experiments}. The author would also like to thank the members of the IBM CHRONOS project~\cite{DBLP:conf/aaai/0001FUA0KHSGBFR24} for their valuable feedback.

%
%
%
\bibliographystyle{splncs04}
\bibliography{references.bib}

\begin{thebibliography}{10}
\providecommand{\url}[1]{\texttt{#1}}
\providecommand{\urlprefix}{URL }
\providecommand{\doi}[1]{https://doi.org/#1}

\bibitem{barnard-mayersAssessingKnowledgeAttitudes2021}
{Barnard-Mayers}, R., Childs, E., Corlin, L., Caniglia, E.C., Fox, M.P.,
  Donnelly, J.P., Murray, E.J.: Assessing knowledge, attitudes, and practices
  towards causal directed acyclic graphs: A qualitative research project.
  European Journal of Epidemiology  \textbf{36}(7),  659--667 (Jul 2021),
  \url{https://doi.org/10.1007/s10654-021-00771-3}

\bibitem{BhandariFHSS21}
Bhandari, M., Feblowitz, M., Hassanzadeh, O., Srinivas, K., Sohrabi, S.:
  Unsupervised causal knowledge extraction from text using natural language
  inference (student abstract). In: {AAAI} (2021)

\bibitem{DBLP:conf/ijcai/BhattacharjyaSH22}
Bhattacharjya, D., Sihag, S., Hassanzadeh, O., Bialik, L.: Summary markov
  models for event sequences. In: Raedt, L.D. (ed.) Proceedings of the
  Thirty-First International Joint Conference on Artificial Intelligence,
  {IJCAI} 2022, Vienna, Austria, 23-29 July 2022. pp. 4836--4842. ijcai.org
  (2022), \url{https://doi.org/10.24963/ijcai.2022/670}

\bibitem{bromiley2015enterprise}
Bromiley, P., McShane, M., Nair, A., Rustambekov, E.: Enterprise risk
  management: Review, critique, and research directions. Long range planning
  \textbf{48}(4),  265--276 (2015)

\bibitem{caselli-vossen-2017-event-storyline}
Caselli, T., Vossen, P.: The event {S}tory{L}ine corpus: A new benchmark for
  causal and temporal relation extraction. In: Proceedings of the Events and
  Stories in the News Workshop. pp. 77--86. Association for Computational
  Linguistics, Vancouver, Canada (Aug 2017). \doi{10.18653/v1/W17-2711},
  \url{https://aclanthology.org/W17-2711}

\bibitem{DBLP:conf/aaai/0001FUA0KHSGBFR24}
Chang, M., Fokoue, A., Uceda{-}Sosa, R., Awasthy, P., Barker, K., Kumaravel,
  S., Hassanzadeh, O., Soares, E.F.S., Gao, T., Bhattacharjya, D., Florian, R.,
  Roukos, S.: {CHRONOS:} {A} schema-based event understanding and prediction
  system. In: Thirty-Eighth {AAAI} Conference on Artificial Intelligence,
  {AAAI} 2024, Thirty-Sixth Conference on Innovative Applications of Artificial
  Intelligence, {IAAI} 2024, Fourteenth Symposium on Educational Advances in
  Artificial Intelligence, {EAAI} 2014, February 20-27, 2024, Vancouver,
  Canada. pp. 22871--22877. {AAAI} Press (2024).
  \doi{10.1609/AAAI.V38I21.30323},
  \url{https://doi.org/10.1609/aaai.v38i21.30323}

\bibitem{tacl/DunietzLC17}
Dunietz, J., Levin, L.S., Carbonell, J.G.: Automatically tagging constructions
  of causation and their slot-fillers. Trans. Assoc. Comput. Linguistics
  \textbf{5},  117--133 (2017)

\bibitem{DunietzLC17}
Dunietz, J., Levin, L.S., Carbonell, J.G.: The {BECauSE} corpus 2.0: Annotating
  causality and overlapping relations. In: Proceedings of the 11th Linguistic
  Annotation Workshop, LAW@EACL. pp. 95--104 (2017),
  \url{https://doi.org/10.18653/v1/W17-0812}

\bibitem{DBLP:conf/semweb/Hassanzadeh21a}
Hassanzadeh, O.: Building a knowledge graph of events and consequences using
  wikidata. In: Proceedings of the 2nd Wikidata Workshop (Wikidata 2021)
  co-located with the 20th International Semantic Web Conference {(ISWC} 2021),
  Virtual Conference, October 24, 2021. {CEUR} Workshop Proceedings, vol.~2982.
  CEUR-WS.org (2021), \url{https://ceur-ws.org/Vol-2982/paper-12.pdf}

\bibitem{ijcai2022p850neatToolkit}
Hassanzadeh, O., Awasthy, P., Barker, K., Bhardwaj, O., Bhattacharjya, D.,
  Feblowitz, M., Martie, L., Ni, J., Srinivas, K., Yip, L.: Knowledge-based
  news event analysis and forecasting toolkit. In: Raedt, L.D. (ed.)
  Proceedings of the Thirty-First International Joint Conference on Artificial
  Intelligence, {IJCAI-22}. pp. 5904--5907. International Joint Conferences on
  Artificial Intelligence Organization (7 2022). \doi{10.24963/ijcai.2022/850},
  \url{https://doi.org/10.24963/ijcai.2022/850}, demo Track

\bibitem{HassanzadehBFSP20}
Hassanzadeh, O., Bhattacharjya, D., Feblowitz, M., Srinivas, K., Perrone, M.,
  Sohrabi, S., Katz, M.: Causal knowledge extraction through large-scale text
  mining. In: {AAAI}. pp. 13610--13611 (2020)

\bibitem{he2023annollm}
He, X., Lin, Z., Gong, Y., Jin, A.L., Zhang, H., Lin, C., Jiao, J., Yiu, S.M.,
  Duan, N., Chen, W.: Annollm: Making large language models to be better
  crowdsourced annotators (2023)

\bibitem{heindorf2020causenet}
Heindorf, S., Scholten, Y., Wachsmuth, H., Ngomo, A.C.N., Potthast, M.:
  {CauseNet}: Towards a causality graph extracted from the web. In: {CIKM}.
  {ACM} (2020)

\bibitem{hendrickx-etal-2010-semeval}
Hendrickx, I., Kim, S.N., Kozareva, Z., Nakov, P., {\'O}~S{\'e}aghdha, D.,
  Pad{\'o}, S., Pennacchiotti, M., Romano, L., Szpakowicz, S.: {S}em{E}val-2010
  task 8: Multi-way classification of semantic relations between pairs of
  nominals. In: Proceedings of the 5th International Workshop on Semantic
  Evaluation. pp. 33--38. Association for Computational Linguistics, Uppsala,
  Sweden (Jul 2010), \url{https://aclanthology.org/S10-1006}

\bibitem{hosseiniCRESTCausalRelation2023}
Hosseini, P.: {{CREST}}: {{A Causal Relation Schema}} for {{Text}} (May 2023),
  \url{https://github.com/phosseini/CREST}

\bibitem{HurriyetogluYMD21}
H{\"{u}}rriyetoglu, A., Y{\"{o}}r{\"{u}}k, E., Mutlu, O., Durusan, F., Yoltar,
  {\c{C}}., Y{\"{u}}ret, D., G{\"{u}}rel, B.: Cross-context news corpus for
  protest event-related knowledge base construction. Data Intell.
  \textbf{3}(2),  308--335 (2021). \doi{10.1162/dint\_a\_00092},
  \url{https://doi.org/10.1162/dint\_a\_00092}

\bibitem{iwamaHOPEGraphHypothesisEvaluation2021}
Iwama, F., Enoki, M., Yoshihama, S.: {{HOPE-Graph}}: {{A Hypothesis Evaluation
  Service}} considering {{News}} and {{Causality Knowledge}}. In: 2021 {{IEEE
  International Conference}} on {{Smart Data Services}} ({{SMDS}}). pp.
  198--209 (Sep 2021). \doi{10.1109/SMDS53860.2021.00034}

\bibitem{jaiminiCausalKGCausalKnowledge2022a}
Jaimini, U., Sheth, A.: {{CausalKG}}: {{Causal Knowledge Graph Explainability
  Using Interventional}} and {{Counterfactual Reasoning}}. IEEE Internet
  Computing  \textbf{26}(1),  43--50 (Jan 2022). \doi{10.1109/MIC.2021.3133551}

\bibitem{DBLP:conf/semweb/KhatiwadaSSH22}
Khatiwada, A., Shirai, S., Srinivas, K., Hassanzadeh, O.: Knowledge graph
  embeddings for causal relation prediction. In: Proceedings of the Workshop on
  Deep Learning for Knowledge Graphs {(DL4KG} 2022) co-located with the 21th
  International Semantic Web Conference {(ISWC} 2022), Virtual Conference,
  online, October 24, 2022. {CEUR} Workshop Proceedings, vol.~3342. CEUR-WS.org
  (2022), \url{https://ceur-ws.org/Vol-3342/paper-8.pdf}

\bibitem{kratzwald-etal-2020-intkb}
Kratzwald, B., Kunpeng, G., Feuerriegel, S., Diefenbach, D.: {I}nt{KB}: A
  verifiable interactive framework for knowledge base completion. In:
  Proceedings of the 28th International Conference on Computational Linguistics
  (COLING) (2020)

\bibitem{lewis2020retrieval}
Lewis, P., Perez, E., Piktus, A., Petroni, F., Karpukhin, V., Goyal, N.,
  K{\"u}ttler, H., Lewis, M., Yih, W.t., Rockt{\"a}schel, T., et~al.:
  Retrieval-augmented generation for knowledge-intensive nlp tasks. Advances in
  Neural Information Processing Systems  \textbf{33},  9459--9474 (2020)

\bibitem{LuoSZHW16CausalNet}
Luo, Z., Sha, Y., Zhu, K.Q., Hwang, S., Wang, Z.: Commonsense causal reasoning
  between short texts. In: Proceedings of the Fifteenth International
  Conference on Principles of Knowledge Representation and Reasoning ({KR})
  (2016)

\bibitem{mariko-etal-2020-financial}
Mariko, D., Abi-Akl, H., Labidurie, E., Durfort, S., De~Mazancourt, H., El-Haj,
  M.: The financial document causality detection shared task ({F}in{C}ausal
  2020). In: Proceedings of the 1st Joint Workshop on Financial Narrative
  Processing and MultiLing Financial Summarisation. pp. 23--32. COLING,
  Barcelona, Spain (Online) (Dec 2020),
  \url{https://aclanthology.org/2020.fnp-1.3}

\bibitem{mariko-etal-2021-financial}
Mariko, D., Akl, H.A., Labidurie, E., Durfort, S., de~Mazancourt, H., El-Haj,
  M.: The financial document causality detection shared task ({F}in{C}ausal
  2021). In: Proceedings of the 3rd Financial Narrative Processing Workshop.
  pp. 58--60. Association for Computational Linguistics, Lancaster, United
  Kingdom (15-16 Sep 2021), \url{https://aclanthology.org/2021.fnp-1.10}

\bibitem{mintz2009distant}
Mintz, M., Bills, S., Snow, R., Jurafsky, D.: Distant supervision for relation
  extraction without labeled data. In: Proceedings of the Joint Conference of
  the 47th Annual Meeting of the ACL and the 4th International Joint Conference
  on Natural Language Processing of the AFNLP. pp. 1003--1011 (2009)

\bibitem{mirza-etal-2014-annotating-causal-timebank}
Mirza, P., Sprugnoli, R., Tonelli, S., Speranza, M.: Annotating causality in
  the {T}emp{E}val-3 corpus. In: Proceedings of the {EACL} 2014 Workshop on
  Computational Approaches to Causality in Language ({CA}to{CL}). pp. 10--19.
  Association for Computational Linguistics, Gothenburg, Sweden (Apr 2014).
  \doi{10.3115/v1/W14-0702}, \url{https://aclanthology.org/W14-0702}

\bibitem{SAGE:ijcai/MorstatterGSBAM19}
Morstatter, F., Galstyan, A., Satyukov, G., Benjamin, D., Abeliuk, A.,
  Mirtaheri, M., Hossain, K.S.M.T., Szekely, P.A., Ferrara, E., Matsui, A.,
  Steyvers, M., Bennett, S., Budescu, D.V., Himmelstein, M., Ward, M.D., Beger,
  A., Catasta, M., Sosic, R., Leskovec, J., Atanasov, P., Joseph, R., Sethi,
  R., Abbas, A.E.: {SAGE:} {A} hybrid geopolitical event forecasting system.
  In: Kraus, S. (ed.) Proceedings of the Twenty-Eighth International Joint
  Conference on Artificial Intelligence, {IJCAI} 2019, Macao, China, August
  10-16, 2019. pp. 6557--6559. ijcai.org (2019). \doi{10.24963/ijcai.2019/955},
  \url{https://doi.org/10.24963/ijcai.2019/955}

\bibitem{embersAt4Years}
Muthiah, S., et~al.: Embers at 4 years: Experiences operating an open source
  indicators forecasting system. In: KDD. pp. 205--214 (2016).
  \doi{10.1145/2939672.2939709}

\bibitem{pearlCausality2009}
Pearl, J.: Causality. {Cambridge University Press} (Sep 2009)

\bibitem{pearlCausalInference2010}
Pearl, J.: Causal {{Inference}}. In: Proceedings of {{Workshop}} on
  {{Causality}}: {{Objectives}} and {{Assessment}} at {{NIPS}} 2008. pp.
  39--58. {PMLR} (Feb 2010),
  \url{https://proceedings.mlr.press/v6/pearl10a.html}

\bibitem{prasadrashmiPennDiscourseTreebank2019}
Prasad, R., Webber, B., Lee, A., Joshi, A.: Penn {{Discourse Treebank Version}}
  3.0 (Mar 2019). \doi{10.35111/QEBF-GK47},
  \url{https://catalog.ldc.upenn.edu/LDC2019T05}

\bibitem{prosperi2020causal}
Prosperi, M., Guo, Y., Sperrin, M., Koopman, J.S., Min, J.S., He, X., Rich, S.,
  Wang, M., Buchan, I.E., Bian, J.: Causal inference and counterfactual
  prediction in machine learning for actionable healthcare. Nature Machine
  Intelligence  \textbf{2}(7),  369--375 (2020)

\bibitem{RadinskyDM12_WWW}
Radinsky, K., Davidovich, S., Markovitch, S.: Learning causality for news
  events prediction. In: WWW (2012)

\bibitem{squad2}
Rajpurkar, P., Jia, R., Liang, P.: Know what you don{'}t know: Unanswerable
  questions for {SQ}u{AD}. In: Proceedings of the 56th Annual Meeting of the
  Association for Computational Linguistics (Volume 2: Short Papers). pp.
  784--789. Association for Computational Linguistics, Melbourne, Australia
  (Jul 2018). \doi{10.18653/v1/P18-2124},
  \url{https://www.aclweb.org/anthology/P18-2124}

\bibitem{DBLP:conf/case-ws/SahaGNHYS22CausalNews}
Saha, A., Gittens, A., Ni, J., Hassanzadeh, O., Yener, B., Srinivas, K.:
  {SPOCK} @ causal news corpus 2022: Cause-effect-signal span detection using
  span-based and sequence tagging models. In: H{\"{u}}rriyetoglu, A., Tanev,
  H., Zavarella, V., Y{\"{o}}r{\"{u}}k, E. (eds.) Proceedings of the 5th
  Workshop on Challenges and Applications of Automated Extraction of
  Socio-political Events from Text, CASE@EMNLP 2022, Abu Dhabi, United Arab
  Emirates (Hybrid), December 7-8, 2022. pp. 133--137. Association for
  Computational Linguistics (2022),
  \url{https://aclanthology.org/2022.case-1.18}

\bibitem{DBLP:journals/corr/abs-1910-01108DistilBERT}
Sanh, V., Debut, L., Chaumond, J., Wolf, T.: {DistilBERT}, a distilled version
  of {BERT:} smaller, faster, cheaper and lighter. CoRR
  \textbf{abs/1910.01108} (2019), \url{http://arxiv.org/abs/1910.01108}

\bibitem{ATOMIC}
Sap, M., LeBras, R., Allaway, E., Bhagavatula, C., Lourie, N., Rashkin, H.,
  Roof, B., Smith, N.A., Choi, Y.: {ATOMIC}: An atlas of machine commonsense
  for if-then reasoning. CoRR  \textbf{abs/1811.00146} (2018),
  \url{http://arxiv.org/abs/1811.00146}

\bibitem{DBLP:conf/www/ShiraiBH23}
Shirai, S., Bhattacharjya, D., Hassanzadeh, O.: Event prediction using
  case-based reasoning over knowledge graphs. In: Ding, Y., Tang, J., Sequeda,
  J.F., Aroyo, L., Castillo, C., Houben, G. (eds.) Proceedings of the {ACM} Web
  Conference 2023, {WWW} 2023, Austin, TX, USA, 30 April 2023 - 4 May 2023. pp.
  2383--2391. {ACM} (2023). \doi{10.1145/3543507.3583201},
  \url{https://doi.org/10.1145/3543507.3583201}

\bibitem{ShiraiBH23}
Shirai, S., Bhattacharjya, D., Hassanzadeh, O.: Event prediction using
  case-based reasoning over knowledge graphs. In: Proceedings of the {ACM} Web
  Conference 2023, {WWW} 2023, Austin, TX, USA, 30 April 2023 - 4 May 2023. pp.
  2383--2391. {ACM} (2023), \url{https://doi.org/10.1145/3543507.3583201}

\bibitem{Sohrabi19}
Sohrabi, S., Katz, M., Hassanzadeh, O., Udrea, O., Feblowitz, M.D., Riabov, A.:
  {IBM} scenario planning advisor: Plan recognition as {AI} planning in
  practice. {AI} Commun.  \textbf{32}(1),  1--13 (2019),
  \url{https://doi.org/10.3233/AIC-180602}

\bibitem{DBLP:conf/aaai/SpeerCH17ConceptNet}
Speer, R., Chin, J., Havasi, C.: Conceptnet 5.5: An open multilingual graph of
  general knowledge. In: Singh, S., Markovitch, S. (eds.) Proceedings of the
  Thirty-First {AAAI} Conference on Artificial Intelligence, February 4-9,
  2017, San Francisco, California, {USA}. pp. 4444--4451. {AAAI} Press (2017),
  \url{http://aaai.org/ocs/index.php/AAAI/AAAI17/paper/view/14972}

\bibitem{tan-etal-2022-event}
Tan, F.A., Hettiarachchi, H., H{\"u}rriyeto{\u{g}}lu, A., Caselli, T., Uca, O.,
  Liza, F.F., Oostdijk, N.: Event causality identification with causal news
  corpus - shared task 3, {CASE} 2022. In: Proceedings of the 5th Workshop on
  Challenges and Applications of Automated Extraction of Socio-political Events
  from Text (CASE). pp. 195--208. Association for Computational Linguistics,
  Abu Dhabi, United Arab Emirates (Hybrid) (Dec 2022),
  \url{https://aclanthology.org/2022.case-1.28}

\bibitem{tanUniCausalUnifiedBenchmark2023}
Tan, F.A., Zuo, X., Ng, S.K.: {{UniCausal}}: {{Unified Benchmark}} and
  {{Repository}} for {{Causal Text Mining}} (Apr 2023).
  \doi{10.48550/arXiv.2208.09163}, \url{http://arxiv.org/abs/2208.09163}

\bibitem{VrandecicK14wikidata}
Vrandecic, D., Kr{\"{o}}tzsch, M.: Wikidata: a free collaborative
  knowledgebase. Commun. {ACM}  \textbf{57}(10),  78--85 (2014)

\bibitem{wu2019zero}
Wu, L., Petroni, F., Josifoski, M., Riedel, S., Zettlemoyer, L.: Zero-shot
  entity linking with dense entity retrieval. In: {EMNLP} (2020)

\bibitem{HealthData:IJSWIS.297145}
Yu, H.Q., Reiff-Marganiec, S.: Learning disease causality knowledge from the
  web of health data. International Journal on Semantic Web and Information
  Systems (IJSWIS)  \textbf{18}(1),  1–19 (apr 2022).
  \doi{10.4018/IJSWIS.297145}, \url{https://doi.org/10.4018/IJSWIS.297145}

\bibitem{yuEventLinkingGrounding2023EveLINK}
Yu, X., Yin, W., Gupta, N., Roth, D.: Event {{Linking}}: {{Grounding Event
  Mentions}} to {{Wikipedia}}. In: Proc. of the {{Conference}} of the
  {{European Chapter}} of the {{Association}} for {{Computational Linguistics}}
  ({{EACL}}) (Feb 2023), \url{http://arxiv.org/abs/2112.07888}

\bibitem{zhao-etal-2022-lmturk}
Zhao, M., Mi, F., Wang, Y., Li, M., Jiang, X., Liu, Q., Schuetze, H.: {LMT}urk:
  Few-shot learners as crowdsourcing workers in a language-model-as-a-service
  framework. In: Findings of the Association for Computational Linguistics:
  NAACL 2022. pp. 675--692. Association for Computational Linguistics, Seattle,
  United States (Jul 2022). \doi{10.18653/v1/2022.findings-naacl.51},
  \url{https://aclanthology.org/2022.findings-naacl.51}

\bibitem{zheng2023judging}
Zheng, L., Chiang, W.L., Sheng, Y., Zhuang, S., Wu, Z., Zhuang, Y., Lin, Z.,
  Li, Z., Li, D., Xing, E.P., Zhang, H., Gonzalez, J.E., Stoica, I.: Judging
  llm-as-a-judge with mt-bench and chatbot arena (2023)

\end{thebibliography}
%




\end{document}